\documentclass[10pt,twocolumn,letterpaper]{article}

\usepackage{cvpr}              

\usepackage{array} 
\usepackage[T1]{fontenc}

\definecolor{cvprblue}{rgb}{0.21,0.49,0.74}
\usepackage[pagebackref,pdftex,breaklinks,colorlinks,allcolors=cvprblue]{hyperref}

\def\paperID{16476} 
\def\confName{CVPR}
\def\confYear{2025}

\title{Don't Mesh with Me: Generating Constructive Solid Geometry Instead of Meshes by Fine-Tuning a Code-Generation LLM}

\author{
Maximilian Mews, 
Ansar Aynetdinov, 
Vivian Schiller, 
Peter Eisert, 
Alan Akbik\\[3pt]
Humboldt-Universität zu Berlin\\
}

\begin{document}
\maketitle
\begin{abstract}
While recent advancements in machine learning, such as LLMs, are revolutionizing software development and creative industries, they have had minimal impact on engineers designing mechanical parts, which remains largely a manual process. Existing approaches to generating 3D geometry most commonly use meshes as a 3D representation. While meshes are suitable for assets in video games or animations, they lack sufficient precision and adaptability for mechanical engineering purposes. This paper introduces a novel approach for the generation of 3D geometry that generates surface-based Constructive Solid Geometry (CSG) by leveraging a code-generation LLM. First, we create a dataset of 3D mechanical parts represented as code scripts by converting Boundary Representation geometry (BREP) into CSG-based Python scripts. Second, we create annotations in natural language using GPT-4. The resulting dataset is used to fine-tune a code-generation LLM. The fine-tuned LLM can complete geometries based on positional input and natural language in a plausible way, demonstrating geometric understanding.
\end{abstract}    
\section{Introduction}

\begin{figure}[htbp]
  \centering
  \begin{minipage}{.12\textwidth}
    \centering
    \includegraphics[width=\linewidth]{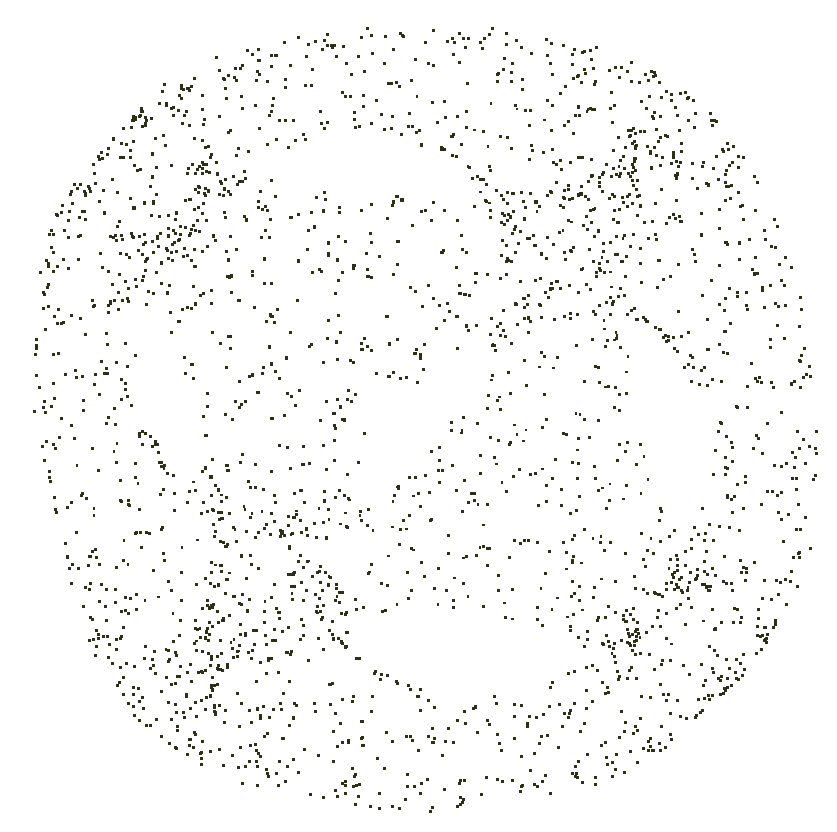}
    \subcaption{Point Cloud}
    \label{fig:4}
  \end{minipage}%
  \begin{minipage}{.12\textwidth}
    \centering
    \includegraphics[width=\linewidth]{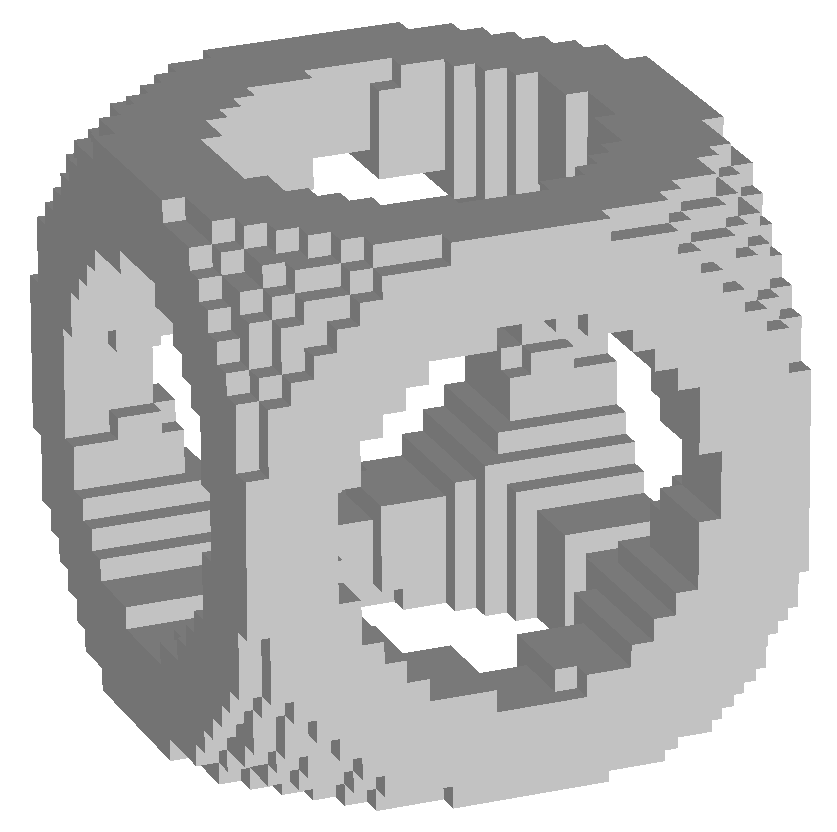}
    \subcaption{Voxel Grid}
    \label{fig:3}
  \end{minipage}%
  \begin{minipage}{.12\textwidth}
    \centering
    \includegraphics[width=\linewidth]{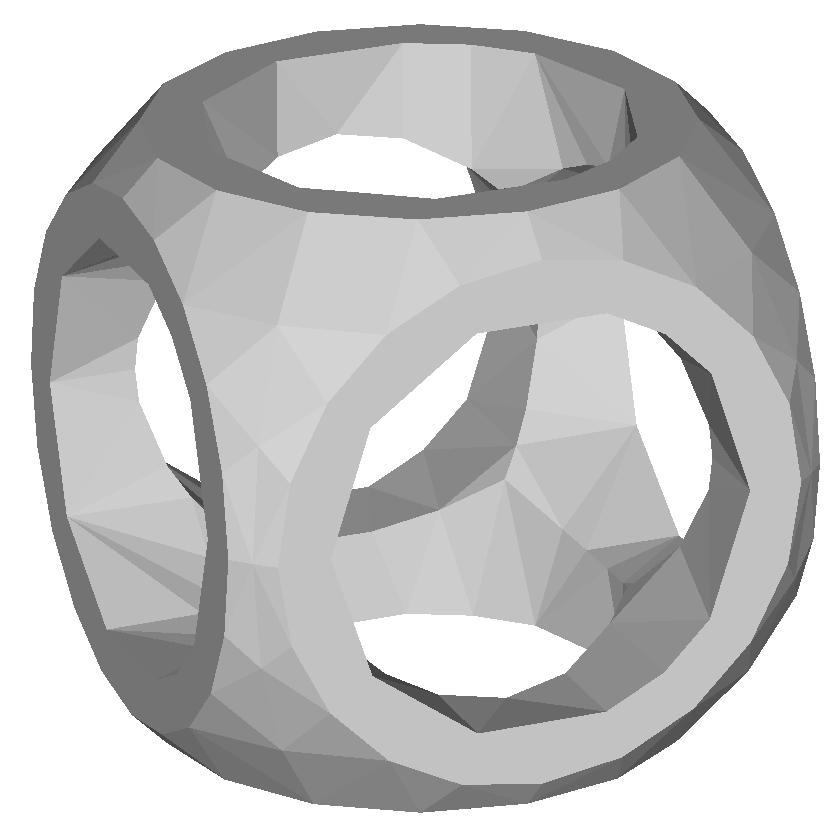}
    \subcaption{Mesh}
    \label{fig:2}
  \end{minipage}%
  \begin{minipage}{.12\textwidth}
    \centering
    \includegraphics[width=\linewidth]{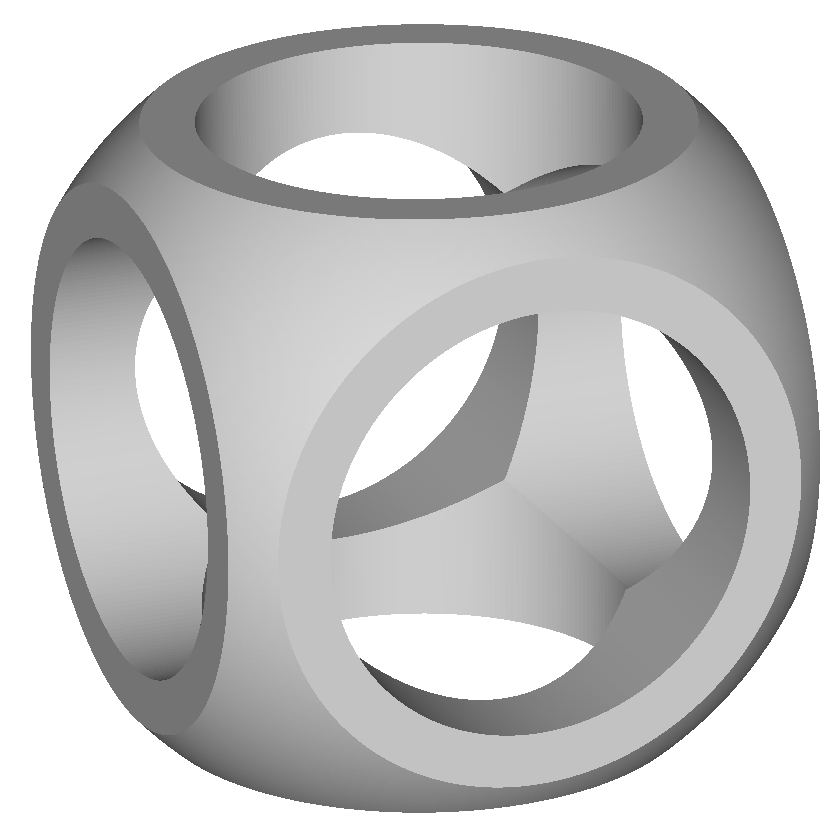}
    \subcaption{CAD/CSG}
    \label{fig:1}
  \end{minipage}
  \caption{Different representations of a 3D geometry.}
  \label{fig:3d-data-representations-reverse}
\end{figure}

In January 2023, ChatGPT \cite{chatgpt} reached 100 million monthly active users just two months after its launch, setting a record for the fastest-growing user base \cite{Hu_2023}. The 2024 Stack Overflow Developer Survey found that 62\% of developers use AI tools in their workflow, with 82\% using them for writing code \cite{StackOverflow2024}. Tools like ChatGPT \cite{chatgpt} and GitHub Copilot \cite{copilot2024} help developers generate or expand code based on text input, speeding up development and extending coding knowledge. Mechanical engineers use CAD software to design 3D parts, but the process remains manual. An AI-driven tool for CAD could accelerate development and allow even non-experts, such as product owners or managers, to create parts through natural language input.

Existing approaches for 3D geometry generation either enable only random generation of mechanical parts instead of text-controlled generation or deliver results in an imprecise format. For instance, approaches such as DreamFusion \cite{poole2022dreamfusiontextto3dusing2d} and MeshGPT \cite{siddiqui2023meshgpt} allow for the generation of 3D geometry based on text input but deliver results in the form of a neural radiance field, which can be converted into a mesh. A mesh is a 3D representation that is made up of a large number of triangles connected by their edges. This means that all non-straight edges and all non-planar surfaces represented by this format are only approximations. This leads to two major issues:
\begin{itemize}
    \item \textbf{Lack of Precision}
    A circle can only be represented as a polygon and a round surface can only be represented as a combination of piecewise planar triangles as shown in \cref{fig:3d-data-representations-reverse}. Therefore a geometry generated by this method is imprecise and lacks smooth edges and surfaces.
    \item \textbf{Lack of modifiability}
    Since all surfaces are represented by a mesh of triangles, the model is not easily modifiable. For example, to change the size of a cylinder, a large number of triangles would need to be correctly identified, then moved and rescaled.
\end{itemize}

Mechanical parts are mostly made of a combination of basic primitives, such as cylinders, cubes or spheres. Directly representing \eg a cylinder by a limited set of parameters such as height, radius and position is way more efficient than approximating it using a triangle mesh. It also gives the model precision and makes modifications easier. This type of 3D representation is called boundary representation (BREP) and is commonly used for CAD software.
DeepCAD \cite{wu2021deepcaddeepgenerativenetwork} is an approach that can generate CAD sequences, however it is able to generate only random mechanical parts and it has no way of controlling the generation via inputs like natural language. 
So far, Text2CAD \cite{khan2024text2cadgeneratingsequentialcad} is the only approach that can generate CAD sequences from text descriptions. This is achieved by training a transformer architecture on a specific dataset of sketch-based CAD sequences created by Onshape \cite{onshape2024} users. This means that the model is limited by data from specific CAD software.

To address the drawbacks of existing approaches, this paper makes the following contributions:

\begin{enumerate}
    \item We propose a dataset creation pipeline \footnote{The code for dataset creation will be made available in our Github repository: \href{https://mxmws.github.io/dont_mesh_with_me}{mxmws.github.io/dont\_mesh\_with\_me}} that takes BREP geometry in the form of .STEP files and converts it into surface-based CSG sequences in the form of Python scripts, in order to allow for model training on 3D files created by any modern CAD software, and not only Onshape.
    \item We produce natural language descriptions of 3D geometries using GPT-4o in order to add text annotations to the training dataset.
    \item We train a code-generation LLM that can complete 3D geometry represented by Python scripts in a plausible way. In addition, it is able to leverage natural language descriptions of said 3D geometries to control the generation of simple geometries.
    
\end{enumerate}

To the best of our knowledge, the resulting model is the first deep learning model specifically trained to generate CSG geometry, and the first LLM that is fine-tuned to complete 3D geometry following text instructions.


    
\section{Related Work}
The field of 3D shape generation has witnessed a diverse array of computational strategies aimed at creating three-dimensional models \cite{
schönhof2022simplified, poole2022dreamfusiontextto3dusing2d, fu2023shapecrafter, siddiqui2023meshgpt}. These strategies predominantly employ geometric representations such as point clouds, voxel grids, or meshes. Despite their widespread use, these representations lack the necessary precision and parametric control that are essential for creating mechanical CAD parts.

The fine-tuning of large language models (LLMs) to enable the generation of 3D geometries remains largely unexplored.

\paragraph{Point-Cloud Generation}

Point cloud generation has progressed through GANs \cite{shu20193dpointcloudgenerative, valsesia2018learning}, invertible flows \cite{yang2019pointflow3dpointcloud}, gradient fields \cite{cai2020learninggradientfieldsshape}, autoencoders \cite{yang2018foldingnetpointcloudautoencoder, zamorski2019adversarialautoencoderscompactrepresentations}, and more recently, diffusion models \cite{zhou20213dshapegenerationcompletion, nichol2022pointegenerating3dpoint, luo2021diffusionprobabilisticmodels3d}, each improving control or representation fidelity.

Point cloud methods excel at handling irregular structures and variable densities, yet their discrete nature compromises surface continuity, hampering high-fidelity representation. Their lack of point connectivity also complicates tasks like surface generation, where structured representations prove advantageous \cite{yang2018foldingnetpointcloudautoencoder}.

\paragraph{Voxel Generation}
Some shape generation methods use Voxel Grids to represent 3D geometry \cite{brock2016generativediscriminativevoxelmodeling,choy20163dr2n2unifiedapproachsingle,wu2017learningprobabilisticlatentspace,tatarchenko2017octreegeneratingnetworksefficient}. A notable advancement in voxel grid generation is the application of autoencoders for model variation within a latent space, as presented by Schönhof \etal \cite{schönhof2022simplified}. By encoding voxel representations into a compact latent space, autoencoders facilitate the generation of diverse voxelized forms through latent space manipulation. Further extending the generative capabilities, 3D-GANs \cite{wu2017learning} (Generative Adversarial Networks) have been employed to produce voxel grids from a learned distribution of shapes.

Voxel graphics can lead to a "blocky" appearance in 3D space, only approximating rather than perfectly representing smooth curves and organic shapes.

\paragraph{Mesh Generation}
The field of generative models has made progress in the creation of mesh-based geometry by manipulating latent spaces \cite{groueix2018atlasnetpapiermacheapproachlearning, chen2020bspnetgeneratingcompactmeshes, nash2020polygenautoregressivegenerativemodel, Hanocka_2019}. Autoencoders have been utilized to enable model variation, as explored by Umetani \cite{10.1145/3145749.3145758}. The field of text-to-mesh generation has seen notable advancements, particularly through approaches that leverage implicit field representations for mesh extraction. Prominent examples include the works of Liu et al.\ \cite{liu2022implicit} and ShapeCrafter \cite{fu2023shapecrafter}, both of which utilize natural language input to guide 3D shape generation. In contrast, MeshGPT \cite{siddiqui2023meshgpt} employs a transformer-based architecture to generate 3D meshes directly, bypassing the need for implicit field modeling. Notably, MeshGPT is capable of producing geometry with sharp and well-defined features.

These methods are adept at creating meshes from text, making them suitable for use in games and animation. However, due to the nature of mesh generation, the resulting models are not able to accurately represent round surfaces, lack parametrization as well as overall precision.

\paragraph{Neural Field Generation}
Neural representations of 3D geometry, referred to as neural implicit fields or neural fields encode the geometry of an object or scene as a continuous function represented by a neural network \cite{mildenhall2020nerfrepresentingscenesneural, park2019deepsdflearningcontinuoussigned}. Building on this foundation, several works have explored using neural fields for generative tasks, such as generating new 3D shapes \cite{genova2019learningshapetemplatesstructured, poole2022dreamfusiontextto3dusing2d, sanghi2022clipforge} or novel views of existing scenes \cite{sitzmann2020scenerepresentationnetworkscontinuous}. These generative neural field approaches offer the potential for flexible and controllable 3D content creation, with applications in areas like computer graphics, virtual reality, and design.

While these approaches excel at representing smooth surfaces and offer continuous representations, they often struggle with high-frequency details, require significant computational resources for training and inference, and can be challenging to edit or manipulate post-generation.

\paragraph{BREP Generation}
BrepGen \cite{xu2024brepgenbrepgenerativediffusion} is a recent diffusion-based method that directly generates CAD models as boundary representations (BREPs). It introduces a hierarchical tree structure where each node encodes geometric primitives, and topology is captured implicitly through node duplication.
BrepGen generates 3D geometry suitable for CAD applications. However, it is not capable of generating geometry based on natural language.

\paragraph{Cuboid Generation}

The ShapeAssembly framework \cite{Jones_2020} operates by writing programs in a domain-specific language named "ShapeAssembly," designed to articulate the assembly of 3D shapes. First, a dataset was created by converting existing 3D models into the ShapeAssembly language, which in turn facilitated the training of a deep generative model, specifically a hierarchical sequence variational autoencoder (VAE).

The scope of ShapeAssembly is confined to the generation of structures composed exclusively of cuboids.

\paragraph{Geometric Code Generation using LLMs}

Research by Makatura \etal \cite{makatura2023large} and Ko \etal \cite{ko2023experimentsgenerativeaipoweredparametric} has explored the potential of pre-trained models such as GPT-3 and GPT-4 to generate coherent 3D geometry in the form of CSG code snippets. 

These models, lacking task-specific fine-tuning, often produce inaccurate geometric outputs even for simple tasks. Their limited spatial reasoning capabilities, stemming from insufficient training data, make precise 3D geometry generation particularly challenging.

\paragraph{CAD Sequence Generation}

Approaches such as DeepCAD \cite{wu2021deepcaddeepgenerativenetwork}, SkexGen \cite{xu2022skexgen} and HNC-CAD \cite{xu2023hierarchicalneuralcodingcontrollable} use transformer based architectures to generate CAD sequences. DeepCAD primarily performs two functions: autoencoding CAD designs and generating random shapes. While their use cases are limited, they showcase the model's ability to produce plausible geometry.

Text2CAD \cite{khan2024text2cadgeneratingsequentialcad} generates CAD sequences from textual descriptions by converting natural language into 2D sketches and 3D operations. The system uses LLMs and VLMs in its data annotation pipeline to create text prompts with varying complexity for CAD models, and employs a transformer-based autoregressive architecture to generate construction sequences. However, the approach depends on specialized CAD sequence training data. In addition it doesn't use LLMs for 3D geometry generation, and cannot complete existing 3D geometries at specified positions.

\paragraph{Constructive Solid Geometry}
Constructive Solid Geometry (CSG) is a 3D modeling technique that uses Boolean operations to combine simple shapes to create more complex ones. These shapes can be geometric primitives like cylinders, cuboids and spheres or cells defined by surfaces. Despite being merely a sequence of simple operations, CSG models can accurately represent complex shapes.

The process of reverse engineering of a CSG sequence from a 3D model involves analyzing the geometric features and structure of the model to decompose it into simpler shapes or surfaces. Several learning based approaches attempt to generate CSG trees from point clouds \cite{NEURIPS2020_63d5fb54, yu2021caprinetlearningcompactcad, ren2021csgstumplearningfriendlycsglike, 10.1145/3321707.3321771}. CSGNet \cite{sharma2018csgnetneuralshapeparser} is also a learning based approach, however it uses a voxel grid as input for the 3D geometry. These learning based approaches are necessary for imprecise input formats such as point clouds and voxel grids, however this also makes them unreliable. McCAD \cite{jne4020031} and GEOUNED \cite{CATALAN20242404} are rule base approaches that take a BREP as input. This enables them to follow the precisely defined geometry of the boundary representations using a set of rules to generate accurate CSG representations in a reliable way. 
\section{Method}

\begin{figure*}[htbp]
  \centering
  \includegraphics[width=0.94\textwidth]{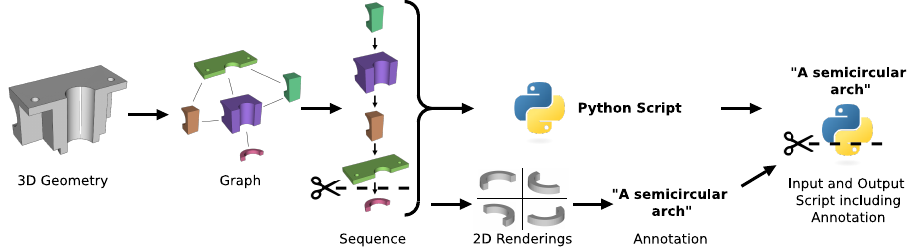}
  \caption{Data generation pipeline. The 3D geometry is first converted into a Graph and then sequentialized. The sequence is split into in- and output and converted into a python script. In parallel the output is rendered as a 2D image and converted into a text annotation which is finally added to the script data.}
  \label{fig:pipeline}
\end{figure*}

\begin{figure}[htbp]
  \centering
  \begin{subfigure}{0.47\textwidth} 
    \centering
    \includegraphics[width=\linewidth]{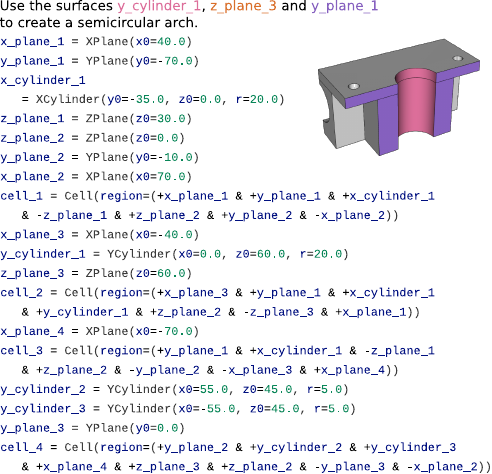}
    \caption{Input}
    \label{fig:input}
  \end{subfigure}
  \begin{subfigure}{0.47\textwidth} 
    \centering
    \includegraphics[width=\linewidth]{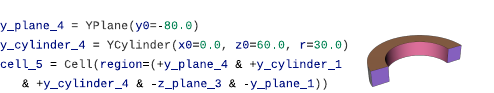}
    \caption{Output}
    \label{fig:output}
  \end{subfigure}
  \caption{Example of a script representing a 3D geometry. The script is split into a base geometry and additional parts that should be learned for geometry completion.}
  \label{fig:code_io}
\end{figure}

\cref{fig:pipeline} and \cref{fig:code_io} show the pipeline for the creation of training data by means of scripts representing the 3D models  and one example for such a script, respectively. The 3D geometry is first converted into a sequence of half-spaces (cells) and split at a specific point which signifies the separation into a subpart geometry and the remaining geometry that leads to the completion of the subpart. Each half-space is defined by a set of hyper-planes and/or cylinders which are represented as reusable variables in a Python script. In parallel to the script generation process, a 2D rendering of the output sequence is created which is then annotated by GPT-4o \cite{openai2023gpt4v}. 

The resulting Python code pair, as shown in \cref{fig:code_io}, consists of the first part of a 3D model as LLM input (\cref{fig:input}) and the second part as LLM target output (\cref{fig:output}). The geometric input and output contain at least one cell each. In addition to the 3D model, the input contains all surface variables that are reused in the output as well as a textual description of the output.

\subsection{Script Generation}\label{section::approach_csg_generation}
Converting the 3D geometries into code scripts requires the following steps: model decomposition, model sequence generation, input-output splitting, input surface determination and the generation of the Python code.

\vspace{-3mm}
\paragraph{Model Decomposition}
CSG generation can be approached through learning-based or rule-based methodologies. Rule-based methods are preferred for their higher reliability but require 3D geometries in a Boundary Representation (BREP) format. Given the focus on mechanical part generation, which commonly utilizes BREP, a rule-based approach is appropriate for this approach. Consequently, GEOUNED \cite{CATALAN20242404}, a tool designed for BREP to CSG conversions, is selected for CSG generation. GEOUNED employs a decomposition process utilizing hyperplanes and non-planar boundaries to create cells (half-spaces) that collectively represent the original 3D geometry. A cell is considered valid, if its volume is finite, meaning it is not open on any of its sides.
The decomposition phase begins with the import of STEP files, followed by their segmentation using GEOUNED.
\vspace{-3mm}
\paragraph{Model Sequence Generation}
A graph of the model is created, visualized by the second image in \cref{fig:pipeline}. This graph contains all half spaces the model is made of as nodes. Two half-spaces are connected in the graph if they touch each other in the model. This graph is used as a basis to generate plausible sequences to generate the same 3D geometry and avoid generating geometry that is unconnected. All plausible sequences are generated to maximize the amount of training data. One plausible sequence is shown in the third image of \cref{fig:pipeline}.
\vspace{-3mm}
\paragraph{Input-Output Splitting}
To prepare the training data, each sequence is divided into model input and target output components. The sequences are split at every possible point between cells to maximize the amount of training data.
\vspace{-3mm}
\paragraph{Input Surface Determination}
For each input-output split, the surfaces required to complete the input are identified. These surfaces are given as a positional input so that the model can learn to generate 3D geometry in a specific location.
\vspace{-3mm}
\paragraph{Code Generation}
In the final step, the sequences created are converted into code. OpenMC \cite{ROMANO201590} is a Python library that supports the creation of hyperplanes and cylinders and their combination into cells. As displayed in \cref{fig:input}, the first line of the input presents the surfaces that need to be reused to create the output geometry. The rest of the script defines hyper-planes and cylinders as variables and uses them to generate cells.

\subsection{Geometry Completion Model}

The geometry completion model serves as a preliminary step towards a text-controlled generative model. It is trained on scripts similar to the one shown in \cref{fig:code_io}, but missing the textual description of the geometry. Its function is to accept an incomplete 3D geometry as input and produce the complete geometry as output by reusing the surfaces given in the input and creating new ones. This is accomplished through the fine-tuning of DeepSeek-Coder-1b \cite{guo2024deepseekcoderlargelanguagemodel}, a code generation model.

\subsection{Annotation}
The text annotation describes the geometry that is to be generated so that the model learns to complete the geometry based on text descriptions. Since the automatic generation of text descriptions from 3D geometry remains under-explored, we use GPT4o for the annotation of 2D images, similar to the Text2CAD approach \cite{khan2024text2cadgeneratingsequentialcad}. The images are simple 2D renderings of the 3D geometry from different perspectives. The following prompt was used to generate text descriptions of the geometric features:
\begin{center}
    \textit{You are shown a part or a set of parts from 4 different angles. Describe the 3D shape of the part(s) in one or more very short informal notes. Do not mention different views. Keep it insanely short. Do not write a full sentence.}
\end{center}
The goal was to keep the annotation as concise as possible and to avoid lengthy explanations usually generated by ChatGPT.

\subsection{Text-Controlled Geometry Completion Model}

The text-controlled geometry completion model is an advancement of the geometry
completion model that utilizes the text annotations as additional training data in order to enable following text instructions describing the features of a 3D model to be generated.

\section{Experiments}

\subsection{Data}\label{sec:data}
\begin{figure}[h!]
  \centering
  \includegraphics[width=\columnwidth]{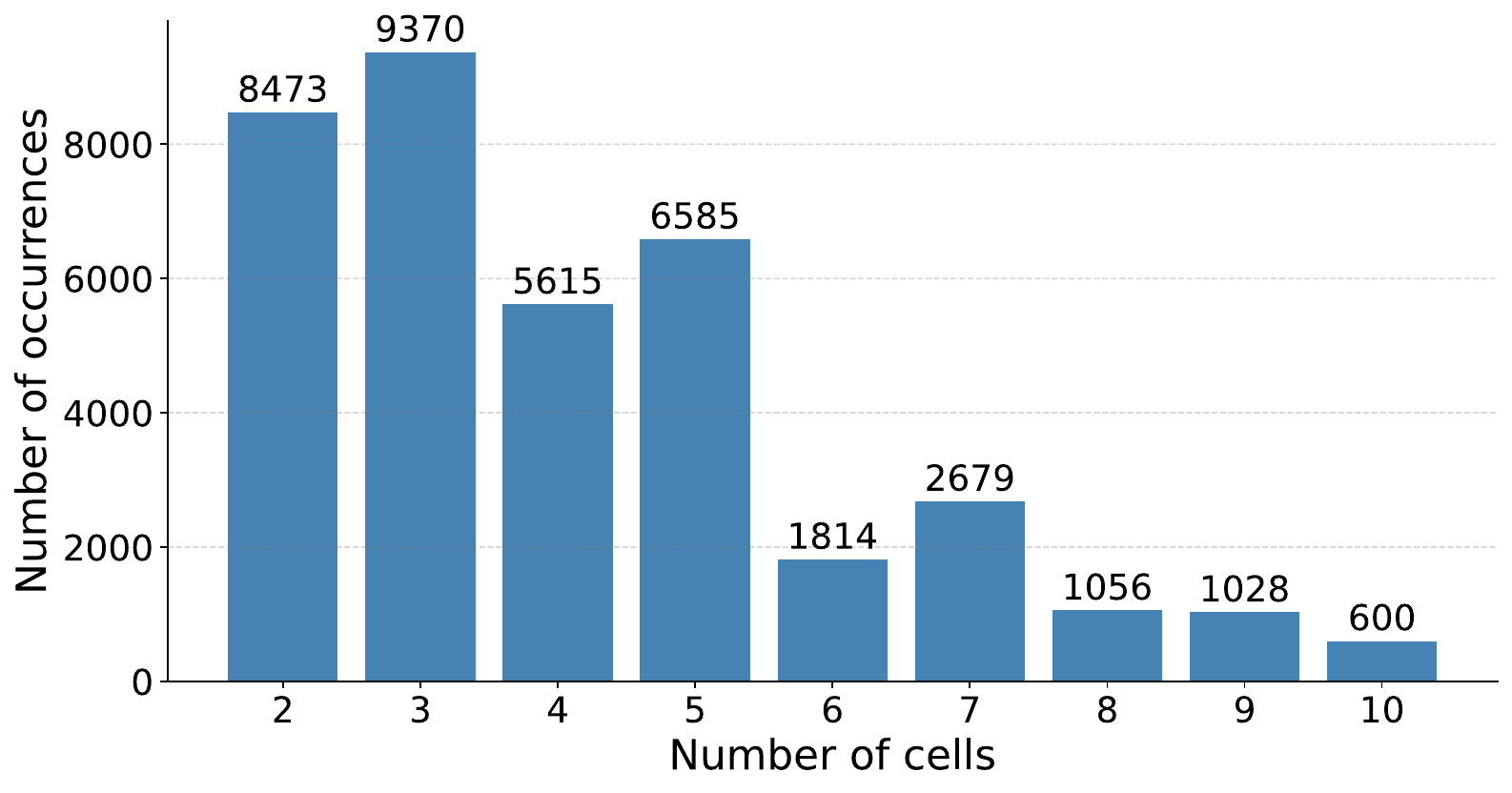}
  \caption{Distribution of parts by number of cells in the curated dataset.}
  \label{fig:parts_by_number_of_cells}
\end{figure}

We chose to evaluate our approach on the ABC dataset \cite{Koch_2019_CVPR}, as in contrast to Fusion360 Gallery \cite{willis2021fusion360gallerydataset} and Thingi10K \cite{zhou2016thingi10kdataset100003dprinting} it is the largest dataset publicly available that contains mechanical parts in a BREP format. Since this work is a proof of concept, the goal is to keep the data as simple as possible. Therefore, only parts were selected from the dataset that can be created by using exclusively hyper-planes and cylinders which are perpendicular or parallel to the axes of the base coordinate system. To keep the sizes of the parts more homogeneous, we removed all parts that contain more than 10 cells. Since we train a completion model, the minimum number of cells needs to be two: one for the input and one for the output. Therefore, we also removed all parts, that only contain 1 cell. The distribution of part sizes is shown in \cref{fig:parts_by_number_of_cells}. In addition, we removed all duplicates from the dataset, including those that differ in scale, rotation or translation. Our resulting dataset contains 37220 parts. With a 90/10 train/test split we have 33498 parts for training. In our experiments, we use two data augmentation methods to increase the size of this dataset. Firstly, we variate where the sequence is cut into input and output, secondly we vary the order of the sequence.

\subsection{Evaluation Strategy}
To the best of our knowledge, no comparable models for the completion of CAD geometries are currently available. DeepCAD \cite{wu2021deepcaddeepgenerativenetwork} and Text2CAD \cite{khan2024text2cadgeneratingsequentialcad} are not able to complete existing geometry in a given position. Therefore, our model is automatically evaluated for plausibility of the geometry generated and manually evaluated for the correctness of the geometry in relation to the natural language input. Additionally, the model is evaluated against a version of itself which doesn't take textual descriptions to prove that textual descriptions increase the likelihood of the model hitting the ground truth.

\subsection{Geometry Completion Model}

The geometry completion model is a simplified version of the final model that does not include text annotations. The purpose of this evaluation is to asses the models ability to generate plausible output geometry based solely on an input geometry.  \cref{tab:model_performance} shows that varying the cuts decreases model performance while varying the order of the sequence increases model performance across all parameters. The parameters that show the syntactic correctness of Python code, connectivity of cells, absence of overlapping cells, combined correctness of syntax and logic, and utilization of all instructed surfaces should ideally be as close to 1.0, as possible. The remaining parameters, displayed in the last three rows of \cref{tab:model_performance}, show the models ability to generate 3D geometries similar or identical to the ground truth, however, generating geometry different from the ground truth is not considered wrong, as there is a near infinite number of correct solutions. Therefore, the absolute value of these parameters is not relevant, however the difference of these parameters for various augmentation strategies is insightful.

\begin{table}[h!]
\centering
\resizebox{0.47\textwidth}{!}{
\begin{tabular}{lcccc}
\hline
\textbf{Variation Method}                  & \textbf{None} & \textbf{Cut} & \textbf{Order} & \textbf{Cut \& Order} \\ \hline
Training Data Size                         & 33,498        & 101,371      & 140,571        & 550,787               \\
Training Time (hours)                      & 2.2           & 6.5          & 9.0            & 35.2                  \\ \hline
Correct Syntax                             & 0.999         & 0.960        & 1.0            & 0.981                 \\
All Cells Connected                        & 0.977         & 0.954        & 0.980          & 0.966                 \\
No Overlapping Cells                       & 0.989         & 0.980        & 0.993          & 0.979                 \\
\textbf{Correct Syntax \& Logic}           & \textbf{0.965}& \textbf{0.895}& \textbf{0.973}& \textbf{0.927}        \\
All Surfaces Used                          & 0.741         & 0.738        & 0.830          & 0.822                 \\
Same Number of Cells                       & 0.494         & 0.439        & 0.580          & 0.565                 \\
Same Syntax, Diff. Params               & 0.105         & 0.100        & 0.159          & 0.167                 \\
Same Syntax, Same Params          & 0.050         & 0.042        & 0.067          & 0.064                 \\ \hline
\end{tabular}}
\caption{Comparison of Model Performance. A geometry is considered logically and syntactically correct when the syntax is correct, all cells are connected and none of them are overlapping.}
\label{tab:model_performance}
\end{table}

In short, \cref{tab:model_performance} shows, that the model is able to generate logically and syntactically correct 3D geometry in the vast majority of cases and augmenting the training data by ordering the sequence differently increases the models performance.

\begin{figure}[h]
  \centering
  \includegraphics[width=0.8\columnwidth]{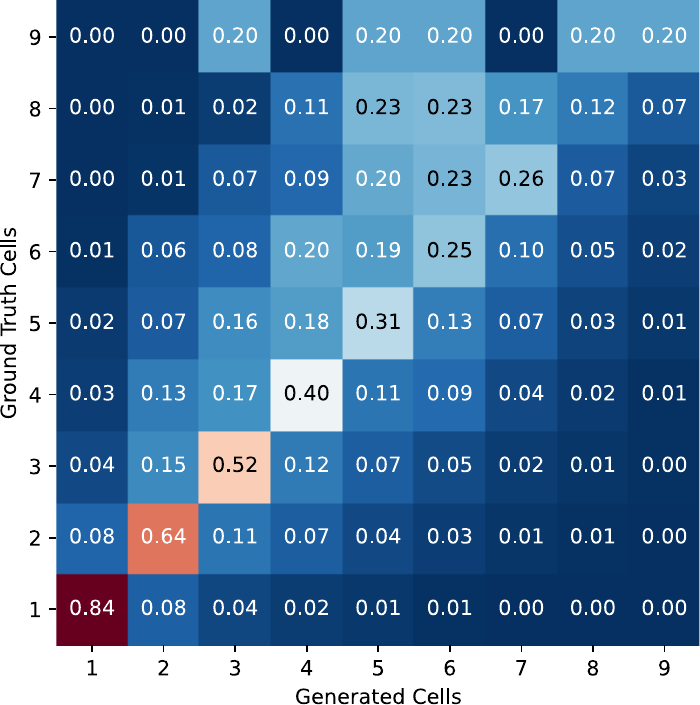}
  \caption{Distribution of the number of cells generated by the geometry completion model for each ground truth cell count. Each row adds up to one and shows the distribution of the number of cells generated for all elements that have a certain number of cells in their ground truth.}
  \label{fig:same_number_of_cells}
\end{figure}

Among other things \cref{tab:model_performance} displays what share of generated geometries have a number of cells equal to the number of cells in the ground truth. \cref{fig:same_number_of_cells} explores this question in more detail. Each row of the matrix represents the distribution of the number of cells generated for a certain number of cells in the corresponding ground truth of the test dataset. For example, the lowest row of the matrix represents all input-output pairs where the ground truth output only consists of one cell. In 84\% of the cases, the actual number of cells generated is also one.

The Figure shows that for almost any number of ground truth cells, the the most common number of cells generated is equal to the number of cells in the ground truth. This indicates that the model understands from the input geometry, as well as the surfaces selected, how complex the resulting geometry should be. With an increasing output cell count the distribution curve gets flatter. This does not mean the model is worse at predicting the cell count of more complex geometries. Naturally, with increasing natural numbers the probability of predicting the correct number decreases, therefore a flatter distribution is to be expected. For the ground truth cell count of 9 the sample size is only 5, which makes it too small to judge the model. Since the total number of cells was cut off at 10 and the input is at least 1 cell, in the training data the target output never had more than 9 cells. In practice the model generated more than 9 cells only in very rare cases.

\begin{figure}[h]
  \centering
  \includegraphics[width=0.8\columnwidth]{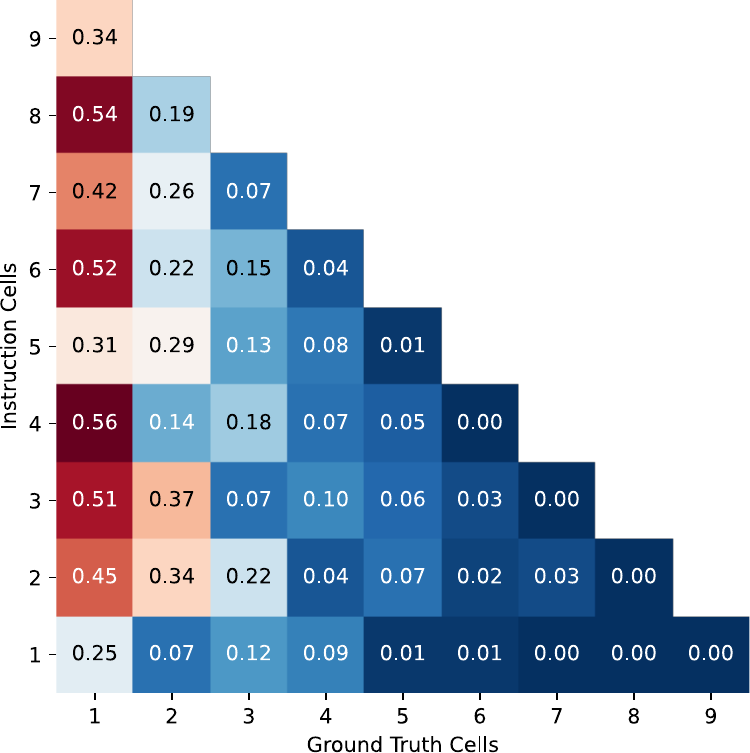}
  \caption{Portion of results where the output generated is equal to the ground truth (differing parameters possible), by number of input cells and number of ground truth cells.}
  \label{fig:equal_to_ground_truth}
\end{figure}

\cref{fig:equal_to_ground_truth} shows the share of results that are syntactically identical to the ground truth in relation to the number of cells in the input and the number of cells in the ground truth output. The parts are only checked for identical structure, meaning a difference in parameters and thus a difference in proportions of the part is possible. For example, the entry in the top row of the first column of the matrix shows, that for the inputs that contained 9 cells and a ground truth output that had 1 cell, 34 percent of them were syntactically identical to the ground truth. A low share of generation results being equal with the output does not mean the model is wrong, as there is an infinite number of possible solutions, not just the output from the ground truth. However, a higher percentage equaling the ground truth indicates the model is better at understanding how to generate structures similar to the ones designed by humans in the test dataset.

The matrix shows that with an increasing number of cells in the ground truth it is less likely that the output generated is equal to the ground truth. This is due to the fact that it is probabilistically less likely to predict a complex structure. The influence of the number of input cells on the accuracy is less clear. It seems like a higher number of input cells increases the chance to generate output equal to the ground truth. A possible explanation for this is that an input with less cells is less specific and therefore leaves more room for further development. If on the other hand an input already contains a specific pattern, the model is likely to continue this pattern.

\subsection{Text annotations}\label{sec:textannotations}

\cref{tab:text_accuracy} shows the accuracy of the text annotations generated. For every number of cells, 100 text annotations were human evaluated and categorized into one of the following categories:

\begin{itemize}
    \item \textbf{Correct}
    The textual description is 100 percent correct and describes the part in its entirety.
    \item \textbf{Partially Correct}
    The textual description is wrong in one aspect or doesn't describe the part in its entirety. Examples of mistakes are wrong counting or mistaking a long thin rectangular beam for a long thin cylinder.
    \item \textbf{Incorrect}
    The textual description is mostly or entirely incorrect.
\end{itemize}

\cref{tab:text_accuracy} shows that the accuracy of the text annotations drops significantly for geometries with more than two cells. Due to an over-representation of geometries with a lower number of cells as displayed in \cref{sec:data}, the average accuracy of the text annotation is 80\%. The fact that the quality of the text annotations is low for complex geometries suggests that a model trained on this data might be worse at generating more complex geometries compared to simple ones.
One of the reasons for the low quality in some of the annotation data is the fact, that some geometries are difficult to describe with words, even by humans. This is especially true when the geometry is complex or is a subpart of another geometry that is difficult to comprehend on its own.

\begin{table}[h!]
\centering
\resizebox{0.47\textwidth}{!}{
\begin{tabular}{cccccccccc}
\hline
\textbf{Number of Cells} & 1 & 2 & 3 & 4 & 5 & 6 & 7 & 8 & 9 \\
\hline
\textbf{Correct} & 0.92 & 0.81 & 0.63 & 0.54 & 0.50 & 0.41 & 0.30 & 0.35 & 0.39 \\
\textbf{Partially Correct} & 0.07 & 0.13 & 0.31 & 0.38 & 0.41 & 0.51 & 0.60 & 0.56 & 0.52 \\
\textbf{Incorrect} & 0.01 & 0.06 & 0.06 & 0.08 & 0.09 & 0.08 & 0.10 & 0.09 & 0.09 \\
\hline
\end{tabular}}
\caption{Accuracy of text annotations, by number of cells.}
\label{tab:text_accuracy}
\end{table}

\subsection{Text-Controlled Geometry Completion Model}

\begin{figure}[htbp]
  \centering
  \includegraphics[width=0.47\textwidth]{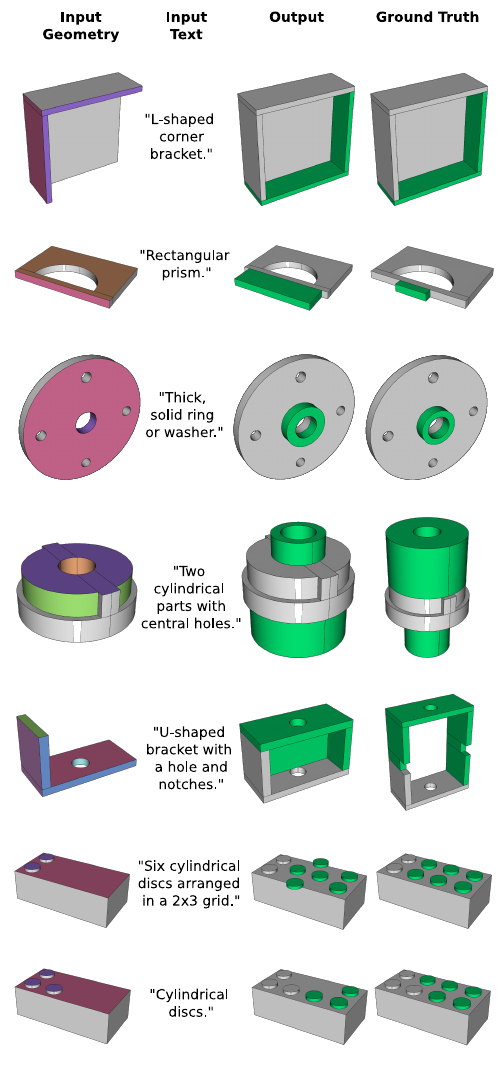}
  \caption{Qualitative results. For the first 4 experiments the model follows the text instructions accurately. For the last 3 experiments the model generates plausible geometry, however it is not aligned with the text instructions.}
  \label{fig:results}
\end{figure}

\cref{tab:gt_compariason} shows the share of generated parts that are equal to the ground truth in relation to the number of cells in the output. Including text annotations of the output in the input increases the likelihood of generating output equal to the ground truth, thus indicating that the output geometry can be controlled via text input. \cref{tab:gt_compariason} indicates that this is true for cell counts of one and two. For outputs with more than two cells the text annotations do not increase the likelihood of generating output equal to the ground truth. \cref{sec:textannotations} shows that the text annotation quality drops significantly for geometries with more than 2 cells. This lack of high quality data explains why the text annotations do not help the model to generate output closer to the ground truth if the output has more than 2 cells.

\begin{table}[h!]
\centering
\resizebox{0.47\textwidth}{!}{
\begin{tabular}{ccccccc}
\hline
\textbf{Ground Truth Cells} & 1 & 2 & 3 & 4 & 5 & 6 \\
\hline
\textbf{Without Text Input} & 0.45 & 0.22 & 0.14 & 0.06 & 0.02 & 0.01 \\
\textbf{With Text Input} & 0.51 & 0.25 & 0.14 & 0.06 & 0.02 & 0.01 \\
\hline
\end{tabular}}
\caption{Comparison of the two models in terms of results where the output generated is equal to the ground truth (differing parameters possible), by number of ground truth cells.}
\label{tab:gt_compariason}
\end{table}

\cref{fig:results} showcases some of the actual generated 3D models. The first column contains renderings of the input geometry including the surfaces that are given as input. The second column shows the text description of the ground truth, which was generated by GPT4o. In columns three and four the ground truth as well as result generated by the model are displayed (green) in combination with the input geometry (gray).

The first row displays an example where the two cells generated are equal to the ones in the ground truth in terms of shape and parameters. The second, third, as well as the fourth row show examples where the ground truth and the geometry generated are equal in terms of shape, but different in terms of parameters. The fifth row displays an example where the model followed the surface input and completed the geometry in a plausible way without adhering to the natural language input. The ground truth consists of six cells, which is a complexity at which the model has difficulty following natural language instructions. 

In the last two rows the input geometries are different incomplete variations of the same toy brick. In the first example the model failed to follow the text description accurately. In the second example the text description was vague and the model followed it accurately, even if it differs from the ground truth. It is notable that oftentimes the model tries to create a symmetric part, presumably because there are many symmetric parts in the training data.

The ability of the model to follow the text instructions is summarized in \cref{tab:text_model_accuracy}. The table was created by evaluating a random sample of 50 examples from the test data for each number of cells manually. The evaluation showed that the accuracy of the model significantly decreases with an increase of the number of cells to be generated. This is due to the low quality of annotation data for more complex geometries, as shown in \cref{sec:textannotations}.

To sum up, the experiments show that the model is able to learn 3D geometry and is able to generate results that are syntactically and mathematically correct and plausible. Furthermore, it is able to follow positional inputs and follow textual inputs as long as the geometry to be generated is not too complex. We hypothesize that with a higher quality of text annotations, the models ability to follow text instructions to generate complex geometries would improve.

\begin{table}[h!]
\centering
\resizebox{0.47\textwidth}{!}{
\begin{tabular}{cccccccc}
\hline
\textbf{Number of Cells} & 1 & 2 & 3 & 4 & 5 & 6 & 7 \\
\hline
\textbf{Correct} & 0.82 & 0.57 & 0.34 & 0.31 & 0.29 & 0.23 & 0.19 \\
\textbf{Partially Correct} & 0.11 & 0.33 & 0.52 & 0.43 & 0.37 & 0.46 & 0.54 \\
\textbf{Incorrect} & 0.07 & 0.10 & 0.14 & 0.26 & 0.34 & 0.31 & 0.27 \\
\hline
\end{tabular}}
\caption{Accuracy of the model following the text instruction, by number of cells.}
\label{tab:text_model_accuracy}
\end{table}

\section{Discussion and Conclusion}

\paragraph{Discussion.}

Our approach demonstrates a promising pipeline for CSG generation, though it faces several limitations. For simplicity, the model was trained exclusively on 3D geometry composed of planes and cylinders oriented perpendicular or parallel to the base coordinate system's axes. Future work could explore generating more advanced geometries, potentially by leveraging a larger code-generation LLM. Additionally, the model's ability to follow text instructions remains limited. One key challenge is that even simple abstract shapes can be remarkably difficult to describe verbally, although paradoxically, complex geometries or operations can sometimes be conveyed with simple words. Therefore, we believe natural language input for CAD will complement rather than replace traditional CAD approaches. For instance, when modifying an existing part, describing the position of modifications verbally is unnecessarily complex, while directly selecting surfaces is more intuitive. While our current model can only expand existing geometry, a modified data pipeline could enable the modification of existing structures. Such modifications are typically straightforward to describe verbally yet time-consuming to implement, such as changing the number of teeth on a gear or adjusting casing dimensions.

\paragraph{Conclusion.}
We introduced a novel approach for generating 3D geometry that leverages a code-generation LLM to produce surface-based Constructive Solid Geometry (CSG). Our pipeline began by converting Boundary Representation geometry (BREP) into CSG-based Python scripts, creating a dataset of 3D mechanical parts represented as code. We then generated natural language annotations using GPT-4 and used this dataset to fine-tune a code-generation LLM. Our approach is both the first deep learning model specifically trained to generate CSG geometry and the first fine-tuned LLM capable of completing 3D geometry based on text instructions. The data pipeline's use of .STEP files as input ensures compatibility across CAD software platforms. Our experiments demonstrate that the model successfully learns 3D geometry, generating results that are syntactically and mathematically sound. Additionally, the model effectively follows both positional and textual inputs for basic geometric structures.

\clearpage
{
    \small
    \bibliographystyle{ieeenat_fullname}
    \bibliography{main}

\begin{thebibliography}{55}
\providecommand{\natexlab}[1]{#1}
\providecommand{\url}[1]{\texttt{#1}}
\expandafter\ifx\csname urlstyle\endcsname\relax
  \providecommand{\doi}[1]{doi: #1}\else
  \providecommand{\doi}{doi: \begingroup \urlstyle{rm}\Url}\fi

\bibitem[Brock et~al.(2016)Brock, Lim, Ritchie, and Weston]{brock2016generativediscriminativevoxelmodeling}
Andrew Brock, Theodore Lim, J.~M. Ritchie, and Nick Weston.
\newblock Generative and discriminative voxel modeling with convolutional neural networks, 2016.

\bibitem[Cai et~al.(2020)Cai, Yang, Averbuch-Elor, Hao, Belongie, Snavely, and Hariharan]{cai2020learninggradientfieldsshape}
Ruojin Cai, Guandao Yang, Hadar Averbuch-Elor, Zekun Hao, Serge Belongie, Noah Snavely, and Bharath Hariharan.
\newblock Learning gradient fields for shape generation, 2020.

\bibitem[Catalán et~al.(2024)Catalán, Sauvan, García, Alguacil, Ogando, and Sanz]{CATALAN20242404}
J.P. Catalán, P. Sauvan, J. García, J. Alguacil, F. Ogando, and J. Sanz.
\newblock Geouned: A new conversion tool from cad to monte carlo geometry.
\newblock \emph{Nuclear Engineering and Technology}, 56\penalty0 (6):\penalty0 2404--2411, 2024.

\bibitem[Chen et~al.(2020)Chen, Tagliasacchi, and Zhang]{chen2020bspnetgeneratingcompactmeshes}
Zhiqin Chen, Andrea Tagliasacchi, and Hao Zhang.
\newblock Bsp-net: Generating compact meshes via binary space partitioning, 2020.

\bibitem[Choy et~al.(2016)Choy, Xu, Gwak, Chen, and Savarese]{choy20163dr2n2unifiedapproachsingle}
Christopher~B. Choy, Danfei Xu, JunYoung Gwak, Kevin Chen, and Silvio Savarese.
\newblock 3d-r2n2: A unified approach for single and multi-view 3d object reconstruction, 2016.

\bibitem[Friedrich et~al.(2019)Friedrich, Fayolle, Gabor, and Linnhoff-Popien]{10.1145/3321707.3321771}
Markus Friedrich, Pierre-Alain Fayolle, Thomas Gabor, and Claudia Linnhoff-Popien.
\newblock Optimizing evolutionary csg tree extraction.
\newblock In \emph{Proceedings of the Genetic and Evolutionary Computation Conference}, page 1183–1191, New York, NY, USA, 2019. Association for Computing Machinery.

\bibitem[Fu et~al.(2023)Fu, Zhan, Chen, Ritchie, and Sridhar]{fu2023shapecrafter}
Rao Fu, Xiao Zhan, Yiwen Chen, Daniel Ritchie, and Srinath Sridhar.
\newblock Shapecrafter: A recursive text-conditioned 3d shape generation model, 2023.

\bibitem[Genova et~al.(2019)Genova, Cole, Vlasic, Sarna, Freeman, and Funkhouser]{genova2019learningshapetemplatesstructured}
Kyle Genova, Forrester Cole, Daniel Vlasic, Aaron Sarna, William~T. Freeman, and Thomas Funkhouser.
\newblock Learning shape templates with structured implicit functions, 2019.

\bibitem[GitHub(2024)]{copilot2024}
GitHub.
\newblock Copilot.
\newblock \url{https://github.com/features/copilot}, 2024.
\newblock Accessed: 2024-08-04.

\bibitem[Groueix et~al.(2018)Groueix, Fisher, Kim, Russell, and Aubry]{groueix2018atlasnetpapiermacheapproachlearning}
Thibault Groueix, Matthew Fisher, Vladimir~G. Kim, Bryan~C. Russell, and Mathieu Aubry.
\newblock Atlasnet: A papier-m\^ach\'e approach to learning 3d surface generation, 2018.

\bibitem[Guo et~al.(2024)Guo, Zhu, Yang, Xie, Dong, Zhang, Chen, Bi, Wu, Li, Luo, Xiong, and Liang]{guo2024deepseekcoderlargelanguagemodel}
Daya Guo, Qihao Zhu, Dejian Yang, Zhenda Xie, Kai Dong, Wentao Zhang, Guanting Chen, Xiao Bi, Y. Wu, Y.~K. Li, Fuli Luo, Yingfei Xiong, and Wenfeng Liang.
\newblock Deepseek-coder: When the large language model meets programming -- the rise of code intelligence, 2024.

\bibitem[Hanocka et~al.(2019)Hanocka, Hertz, Fish, Giryes, Fleishman, and Cohen-Or]{Hanocka_2019}
Rana Hanocka, Amir Hertz, Noa Fish, Raja Giryes, Shachar Fleishman, and Daniel Cohen-Or.
\newblock Meshcnn: a network with an edge.
\newblock \emph{ACM Transactions on Graphics}, 38\penalty0 (4):\penalty0 1–12, 2019.

\bibitem[Harb et~al.(2023)Harb, Leichtle, and Fischer]{jne4020031}
Moataz Harb, Dieter Leichtle, and Ulrich Fischer.
\newblock A novel algorithm for cad to csg conversion in mccad.
\newblock \emph{Journal of Nuclear Engineering}, 4\penalty0 (2):\penalty0 436--447, 2023.

\bibitem[Hu(2023)]{Hu_2023}
Krystal Hu.
\newblock Chatgpt sets record for fastest-growing user base - analyst note | reuters, 2023.

\bibitem[Jones et~al.(2020)Jones, Barton, Xu, Wang, Jiang, Guerrero, Mitra, and Ritchie]{Jones_2020}
R.~Kenny Jones, Theresa Barton, Xianghao Xu, Kai Wang, Ellen Jiang, Paul Guerrero, Niloy~J. Mitra, and Daniel Ritchie.
\newblock Shapeassembly: learning to generate programs for 3d shape structure synthesis.
\newblock \emph{ACM Transactions on Graphics}, 39\penalty0 (6):\penalty0 1–20, 2020.

\bibitem[Kania et~al.(2020)Kania, Zieba, and Kajdanowicz]{NEURIPS2020_63d5fb54}
Kacper Kania, Maciej Zieba, and Tomasz Kajdanowicz.
\newblock Ucsg-net- unsupervised discovering of constructive solid geometry tree.
\newblock In \emph{Advances in Neural Information Processing Systems}, pages 8776--8786. Curran Associates, Inc., 2020.

\bibitem[Khan et~al.(2024)Khan, Sinha, Sheikh, Stricker, Ali, and Afzal]{khan2024text2cadgeneratingsequentialcad}
Mohammad~Sadil Khan, Sankalp Sinha, Talha~Uddin Sheikh, Didier Stricker, Sk~Aziz Ali, and Muhammad~Zeshan Afzal.
\newblock Text2cad: Generating sequential cad models from beginner-to-expert level text prompts, 2024.

\bibitem[Ko et~al.(2023)Ko, Ajibefun, and Yan]{ko2023experimentsgenerativeaipoweredparametric}
Jaechang Ko, John Ajibefun, and Wei Yan.
\newblock Experiments on generative ai-powered parametric modeling and bim for architectural design, 2023.

\bibitem[Koch et~al.(2019)Koch, Matveev, Jiang, Williams, Artemov, Burnaev, Alexa, Zorin, and Panozzo]{Koch_2019_CVPR}
Sebastian Koch, Albert Matveev, Zhongshi Jiang, Francis Williams, Alexey Artemov, Evgeny Burnaev, Marc Alexa, Denis Zorin, and Daniele Panozzo.
\newblock Abc: A big cad model dataset for geometric deep learning.
\newblock In \emph{The IEEE Conference on Computer Vision and Pattern Recognition (CVPR)}, 2019.

\bibitem[Liu et~al.(2022)Liu, Wang, Qi, and Fu]{liu2022implicit}
Zhengzhe Liu, Yi Wang, Xiaojuan Qi, and Chi-Wing Fu.
\newblock Towards implicit text-guided 3d shape generation, 2022.

\bibitem[Luo and Hu(2021)]{luo2021diffusionprobabilisticmodels3d}
Shitong Luo and Wei Hu.
\newblock Diffusion probabilistic models for 3d point cloud generation, 2021.

\bibitem[Makatura et~al.(2023)Makatura, Foshey, Wang, HähnLein, Ma, Deng, Tjandrasuwita, Spielberg, Owens, Chen, Zhao, Zhu, Norton, Gu, Jacob, Li, Schulz, and Matusik]{makatura2023large}
Liane Makatura, Michael Foshey, Bohan Wang, Felix HähnLein, Pingchuan Ma, Bolei Deng, Megan Tjandrasuwita, Andrew Spielberg, Crystal~Elaine Owens, Peter~Yichen Chen, Allan Zhao, Amy Zhu, Wil~J Norton, Edward Gu, Joshua Jacob, Yifei Li, Adriana Schulz, and Wojciech Matusik.
\newblock How can large language models help humans in design and manufacturing?, 2023.

\bibitem[Mildenhall et~al.(2020)Mildenhall, Srinivasan, Tancik, Barron, Ramamoorthi, and Ng]{mildenhall2020nerfrepresentingscenesneural}
Ben Mildenhall, Pratul~P. Srinivasan, Matthew Tancik, Jonathan~T. Barron, Ravi Ramamoorthi, and Ren Ng.
\newblock Nerf: Representing scenes as neural radiance fields for view synthesis, 2020.

\bibitem[Nash et~al.(2020)Nash, Ganin, Eslami, and Battaglia]{nash2020polygenautoregressivegenerativemodel}
Charlie Nash, Yaroslav Ganin, S.~M.~Ali Eslami, and Peter~W. Battaglia.
\newblock Polygen: An autoregressive generative model of 3d meshes, 2020.

\bibitem[Nichol et~al.(2022)Nichol, Jun, Dhariwal, Mishkin, and Chen]{nichol2022pointegenerating3dpoint}
Alex Nichol, Heewoo Jun, Prafulla Dhariwal, Pamela Mishkin, and Mark Chen.
\newblock Point-e: A system for generating 3d point clouds from complex prompts, 2022.

\bibitem[Onshape(n.d.)]{onshape2024}
Onshape.
\newblock Onshape: Cloud-native product development platform.
\newblock \url{https://www.onshape.com}, n.d.
\newblock Accessed: 2024-11-11.

\bibitem[OpenAI(2022)]{chatgpt}
OpenAI.
\newblock Chatgpt: A large-scale generative model for open-domain chat.
\newblock \url{https://openai.com/blog/chatgpt}, 2022.

\bibitem[OpenAI(2023)]{openai2023gpt4v}
OpenAI.
\newblock Gpt-4v(ision) system card, 2023.

\bibitem[Park et~al.(2019)Park, Florence, Straub, Newcombe, and Lovegrove]{park2019deepsdflearningcontinuoussigned}
Jeong~Joon Park, Peter Florence, Julian Straub, Richard Newcombe, and Steven Lovegrove.
\newblock Deepsdf: Learning continuous signed distance functions for shape representation, 2019.

\bibitem[Poole et~al.(2022)Poole, Jain, Barron, and Mildenhall]{poole2022dreamfusiontextto3dusing2d}
Ben Poole, Ajay Jain, Jonathan~T. Barron, and Ben Mildenhall.
\newblock Dreamfusion: Text-to-3d using 2d diffusion, 2022.

\bibitem[Ren et~al.(2021)Ren, Zheng, Cai, Li, Jiang, Cai, Zhang, Pan, Zhang, Zhao, and Yi]{ren2021csgstumplearningfriendlycsglike}
Daxuan Ren, Jianmin Zheng, Jianfei Cai, Jiatong Li, Haiyong Jiang, Zhongang Cai, Junzhe Zhang, Liang Pan, Mingyuan Zhang, Haiyu Zhao, and Shuai Yi.
\newblock Csg-stump: A learning friendly csg-like representation for interpretable shape parsing, 2021.

\bibitem[Romano et~al.(2015)Romano, Horelik, Herman, Nelson, Forget, and Smith]{ROMANO201590}
Paul~K. Romano, Nicholas~E. Horelik, Bryan~R. Herman, Adam~G. Nelson, Benoit Forget, and Kord Smith.
\newblock Openmc: A state-of-the-art monte carlo code for research and development.
\newblock \emph{Annals of Nuclear Energy}, 82:\penalty0 90--97, 2015.
\newblock Joint International Conference on Supercomputing in Nuclear Applications and Monte Carlo 2013, SNA + MC 2013. Pluri- and Trans-disciplinarity, Towards New Modeling and Numerical Simulation Paradigms.

\bibitem[Sanghi et~al.(2022)Sanghi, Chu, Lambourne, Wang, Cheng, Fumero, and Malekshan]{sanghi2022clipforge}
Aditya Sanghi, Hang Chu, Joseph~G. Lambourne, Ye Wang, Chin-Yi Cheng, Marco Fumero, and Kamal~Rahimi Malekshan.
\newblock Clip-forge: Towards zero-shot text-to-shape generation, 2022.

\bibitem[Schönhof et~al.(2022)Schönhof, Elstner, Manea, Tauber, Awad, and Huber]{schönhof2022simplified}
Raoul Schönhof, Jannes Elstner, Radu Manea, Steffen Tauber, Ramez Awad, and Marco~F. Huber.
\newblock Simplified learning of cad features leveraging a deep residual autoencoder, 2022.

\bibitem[Sharma et~al.(2018)Sharma, Goyal, Liu, Kalogerakis, and Maji]{sharma2018csgnetneuralshapeparser}
Gopal Sharma, Rishabh Goyal, Difan Liu, Evangelos Kalogerakis, and Subhransu Maji.
\newblock Csgnet: Neural shape parser for constructive solid geometry, 2018.

\bibitem[Shu et~al.(2019)Shu, Park, and Kwon]{shu20193dpointcloudgenerative}
Dong~Wook Shu, Sung~Woo Park, and Junseok Kwon.
\newblock 3d point cloud generative adversarial network based on tree structured graph convolutions, 2019.

\bibitem[Siddiqui et~al.(2023)Siddiqui, Alliegro, Artemov, Tommasi, Sirigatti, Rosov, Dai, and Nie{\ss}ner]{siddiqui2023meshgpt}
Yawar Siddiqui, Antonio Alliegro, Alexey Artemov, Tatiana Tommasi, Daniele Sirigatti, Vladislav Rosov, Angela Dai, and Matthias Nie{\ss}ner.
\newblock Meshgpt: Generating triangle meshes with decoder-only transformers.
\newblock \emph{arXiv preprint arXiv:2311.15475}, 2023.

\bibitem[Sitzmann et~al.(2020)Sitzmann, Zollhöfer, and Wetzstein]{sitzmann2020scenerepresentationnetworkscontinuous}
Vincent Sitzmann, Michael Zollhöfer, and Gordon Wetzstein.
\newblock Scene representation networks: Continuous 3d-structure-aware neural scene representations, 2020.

\bibitem[{Stack Overflow}(2024)]{StackOverflow2024}
{Stack Overflow}.
\newblock {2024 Stack Overflow Developer Survey}, 2024.

\bibitem[Tatarchenko et~al.(2017)Tatarchenko, Dosovitskiy, and Brox]{tatarchenko2017octreegeneratingnetworksefficient}
Maxim Tatarchenko, Alexey Dosovitskiy, and Thomas Brox.
\newblock Octree generating networks: Efficient convolutional architectures for high-resolution 3d outputs, 2017.

\bibitem[Umetani(2017)]{10.1145/3145749.3145758}
Nobuyuki Umetani.
\newblock Exploring generative 3d shapes using autoencoder networks.
\newblock In \emph{SIGGRAPH Asia 2017 Technical Briefs}, New York, NY, USA, 2017. Association for Computing Machinery.

\bibitem[Valsesia et~al.(2019)Valsesia, Fracastoro, and Magli]{valsesia2018learning}
Diego Valsesia, Giulia Fracastoro, and Enrico Magli.
\newblock Learning localized generative models for 3d point clouds via graph convolution.
\newblock In \emph{International Conference on Learning Representations}, 2019.

\bibitem[Willis et~al.(2021)Willis, Pu, Luo, Chu, Du, Lambourne, Solar-Lezama, and Matusik]{willis2021fusion360gallerydataset}
Karl D.~D. Willis, Yewen Pu, Jieliang Luo, Hang Chu, Tao Du, Joseph~G. Lambourne, Armando Solar-Lezama, and Wojciech Matusik.
\newblock Fusion 360 gallery: A dataset and environment for programmatic cad construction from human design sequences, 2021.

\bibitem[Wu et~al.(2017{\natexlab{a}})Wu, Zhang, Xue, Freeman, and Tenenbaum]{wu2017learning}
Jiajun Wu, Chengkai Zhang, Tianfan Xue, William~T. Freeman, and Joshua~B. Tenenbaum.
\newblock Learning a probabilistic latent space of object shapes via 3d generative-adversarial modeling, 2017{\natexlab{a}}.

\bibitem[Wu et~al.(2017{\natexlab{b}})Wu, Zhang, Xue, Freeman, and Tenenbaum]{wu2017learningprobabilisticlatentspace}
Jiajun Wu, Chengkai Zhang, Tianfan Xue, William~T. Freeman, and Joshua~B. Tenenbaum.
\newblock Learning a probabilistic latent space of object shapes via 3d generative-adversarial modeling, 2017{\natexlab{b}}.

\bibitem[Wu et~al.(2021)Wu, Xiao, and Zheng]{wu2021deepcaddeepgenerativenetwork}
Rundi Wu, Chang Xiao, and Changxi Zheng.
\newblock Deepcad: A deep generative network for computer-aided design models, 2021.

\bibitem[Xu et~al.(2022)Xu, Willis, Lambourne, Cheng, Jayaraman, and Furukawa]{xu2022skexgen}
Xiang Xu, Karl~DD Willis, Joseph~G Lambourne, Chin-Yi Cheng, Pradeep~Kumar Jayaraman, and Yasutaka Furukawa.
\newblock Skexgen: Autoregressive generation of cad construction sequences with disentangled codebooks.
\newblock In \emph{International Conference on Machine Learning}, pages 24698--24724. PMLR, 2022.

\bibitem[Xu et~al.(2023)Xu, Jayaraman, Lambourne, Willis, and Furukawa]{xu2023hierarchicalneuralcodingcontrollable}
Xiang Xu, Pradeep~Kumar Jayaraman, Joseph~G. Lambourne, Karl D.~D. Willis, and Yasutaka Furukawa.
\newblock Hierarchical neural coding for controllable cad model generation, 2023.

\bibitem[Xu et~al.(2024)Xu, Lambourne, Jayaraman, Wang, Willis, and Furukawa]{xu2024brepgenbrepgenerativediffusion}
Xiang Xu, Joseph~G. Lambourne, Pradeep~Kumar Jayaraman, Zhengqing Wang, Karl D.~D. Willis, and Yasutaka Furukawa.
\newblock Brepgen: A b-rep generative diffusion model with structured latent geometry, 2024.

\bibitem[Yang et~al.(2019)Yang, Huang, Hao, Liu, Belongie, and Hariharan]{yang2019pointflow3dpointcloud}
Guandao Yang, Xun Huang, Zekun Hao, Ming-Yu Liu, Serge Belongie, and Bharath Hariharan.
\newblock Pointflow: 3d point cloud generation with continuous normalizing flows, 2019.

\bibitem[Yang et~al.(2018)Yang, Feng, Shen, and Tian]{yang2018foldingnetpointcloudautoencoder}
Yaoqing Yang, Chen Feng, Yiru Shen, and Dong Tian.
\newblock Foldingnet: Point cloud auto-encoder via deep grid deformation, 2018.

\bibitem[Yu et~al.(2021)Yu, Chen, Li, Sanghi, Shayani, Mahdavi-Amiri, and Zhang]{yu2021caprinetlearningcompactcad}
Fenggen Yu, Zhiqin Chen, Manyi Li, Aditya Sanghi, Hooman Shayani, Ali Mahdavi-Amiri, and Hao Zhang.
\newblock Capri-net: Learning compact cad shapes with adaptive primitive assembly, 2021.

\bibitem[Zamorski et~al.(2019)Zamorski, Zięba, Klukowski, Nowak, Kurach, Stokowiec, and Trzciński]{zamorski2019adversarialautoencoderscompactrepresentations}
Maciej Zamorski, Maciej Zięba, Piotr Klukowski, Rafał Nowak, Karol Kurach, Wojciech Stokowiec, and Tomasz Trzciński.
\newblock Adversarial autoencoders for compact representations of 3d point clouds, 2019.

\bibitem[Zhou et~al.(2021)Zhou, Du, and Wu]{zhou20213dshapegenerationcompletion}
Linqi Zhou, Yilun Du, and Jiajun Wu.
\newblock 3d shape generation and completion through point-voxel diffusion, 2021.

\bibitem[Zhou and Jacobson(2016)]{zhou2016thingi10kdataset100003dprinting}
Qingnan Zhou and Alec Jacobson.
\newblock Thingi10k: A dataset of 10,000 3d-printing models, 2016.

\end{thebibliography}
}


\end{document}